\documentclass[runningheads]{llncs}
\usepackage{algpseudocode}
\usepackage{algorithm}

\usepackage{amsmath,amssymb}

\usepackage[T1]{fontenc}
\usepackage{graphicx}
\usepackage{hyperref}
\usepackage{color}

\usepackage{booktabs}

\newcommand{\refalgorithm}[1]{Algorithm~\ref{#1}}

\newcommand{\reffigure}[1]{\figurename~\ref{#1}}

\newcommand{\refsection}[1]{Section~\ref{#1}}
\newcommand{\refsubsection}[1]{Subsection~\ref{#1}}
\newcommand{\reftable}[1]{Table~\ref{#1}}

\newcommand{\argmax}{\operatornamewithlimits{argmax}}

\begin{document}
\title{Anytime Sequential Halving in\\ Monte-Carlo Tree Search}
%
%
\author{Dominic Sagers
\and Mark H.M. Winands
\and Dennis J.N.J. Soemers
}
\authorrunning{D. Sagers et al.}
%
\institute{Department of Advanced Computing Sciences, Maastricht University,\\ Paul-Henri Spaaklaan 1, 6229 EN Maastricht, the Netherlands
\email{dominicsagers1@gmail.com}, \email{\{m.winands,dennis.soemers\}@maastrichtuniversity.nl}}
\maketitle              
\begin{abstract}
Monte-Carlo Tree Search (MCTS) typically uses multi-armed bandit (MAB) strategies designed to minimize cumulative regret, such as UCB1, as its selection strategy. However, in the root node of the search tree, it is more sensible to minimize simple regret. Previous work has proposed using Sequential Halving as selection strategy in the root node, as, in theory, it performs better with respect to simple regret. However, Sequential Halving requires a budget of iterations to be predetermined, which is often impractical. This paper proposes an \emph{anytime} version of the algorithm, which can be halted at any arbitrary time and still return a satisfactory result, while being designed such that it approximates the behavior of Sequential Halving. Empirical results in synthetic MAB problems and ten different board games demonstrate that the algorithm's performance is competitive with Sequential Halving and UCB1 (and their analogues in MCTS).

\end{abstract}
\section{Introduction}

 Monte-Carlo Tree Search (MCTS) \cite{Kocsis_2006_Bandit,Coulom_2007_MCTS} is a search algorithm used for different sequential decision-making problems. It has been thoroughly studied within the context of game playing agents, but also seen use in other planning, optimization, and control problems \cite{Browne_2012_MCTS}. The algorithm consists of four strategic steps, each of which can be implemented using a variety of different strategies \cite{Browne_2012_MCTS,Swiechowski_2022_MCTS}. Strategies for the \textit{selection} step tend to use Multi-Armed Bandit (MAB) algorithms, which balance exploration (sampling actions that are less explored) with exploitation (more deeply searching actions that appear more promising).
%

The most commonly used selection strategy is UCB1 \cite{Auer_2002_Finite}, which focuses on minimizing \textit{cumulative regret}. Sequential Halving (SH) \cite{Karnin_2013_Almost}, which focuses on \textit{simple regret}, may be argued to be a more suitable choice in MCTS \cite{Pepels_2014_Minimizing,Cazenave_2015_SHOT}.
Integrations of SH into MCTS have been described in research on partially observable games \cite{Pepels_2016_Sequential}, variants of MCTS that take additional guidance from scores learned through online or offline learning \cite{Fabiano_2022_Sequential}, and the state-of-the-art Gumbel AlphaZero and MuZero \cite{Danihelka_2022_Policy}, which also integrate deep neural networks.

Running through the four strategic steps of MCTS once is referred to as an iteration, and MCTS typically runs multiple iterations, after which it returns a final decision (e.g., move to play or action to take). It is common to use either a time budget, where MCTS keeps running iterations until it runs out of time, or an iteration budget, where it runs a predetermined number of iterations. SH requires the number of iterations that can be executed to be known in advance, which means that MCTS using SH as selection strategy does not have the \textit{anytime} property. The algorithm cannot be terminated at any arbitrary point in time and be expected to have the quality of its final decision smoothly increasing as processing time increases (barring pathological cases \cite{Nguyen_2024_Lookahead}). When dealing with known games (for which the average number of iterations per unit of time could be measured) and a fixed per-move time budgets, a reasonable approximation of a fixed iteration budget may be calculated and used. However, when dealing with particularly large and varied sets of games \cite{Soemers_2024_MCTSDataset}, automatically generated games that must be played quickly in the context of an evolutionary search \cite{Todd_2024_GAVEL}, or agents that automatically manage their time in an intelligent manner \cite{Huang_2010_TimeManagement,Baier_2015_TimeManagement}, the lack of anytime property can be more problematic.

This paper proposes \textit{anytime SH}: a MAB algorithm with the anytime property, which can be used as selection strategy in (the root node of) MCTS. Its design was heavily inspired by the original SH, with anytime SH essentially being the original SH turned inside out. While we leave formal analyses of bounds on regret for future work, empirical results in synthetic MAB problems as well as a diverse set of ten board games demonstrate that anytime SH performs competitively with UCB1 (or UCT in games) as well as SH (only used in root node in games) in practice, whilst---in contrast to SH---retaining the anytime property.

\section{Background} \label{Sec:Background}

The Multi-armed Bandit (MAB) problem assumes an environment in which an agent repeatedly chooses one from a set of $K$ arms to pull at time steps $t = 1, 2, \dots$. Each of the arms is associated with a stationary, unknown probability distribution, and whenever the agent pulls an arm with index $i_t$ at time $t$, it receives a reward sampled from the corresponding distribution with (unknown) mean $\mu_{i_t}$. The most common measure of performance in MAB problems is the \textit{cumulative regret} $R(T) = \sum_{t=1}^T \left( \mu^* - \mu_{i_t} \right)$, which accumulates the regret of not having always picked the optimal arm (with the highest mean $\mu^*$) over a sequence of $T$ time steps. The \textit{simple regret} $r(T) = \mu^* - \mu_{i_{T+1}}$, which solely measures the regret of a single choice made after a period of $T$ exploration steps, is an alternative measure of performance \cite{Audibert_2010_BestArm,Bubeck_2011_Pure,Feldman_2014_Simple}. Simple regret is a more appropriate measure of performance when it is only an ultimate single decision that matters, with all prior decisions simply serving as a learning or training phase.

UCB1 \cite{Auer_2002_Finite} is a common MAB algorithm, used as the selection strategy in the canonical UCT \cite{Kocsis_2006_Bandit} variant of MCTS. It is designed primarily to minimize cumulative regret.
At any give time step $t$, UCB1 selects an arm $j_t$ to pull using $j_t = \argmax_{1 \leq j \leq K} X_j + C \sqrt{\frac{\ln \left( \sum_{k=1}^K n_k \right)}{n_j}}$, where $X_j$ is the average reward collected from previous pulls of arm $j$ so far, $C$ is a tunable hyperparameter, which affects the balance between exploration and exploitation, and $n_i$ is the number of times an arm $i$ has been pulled so far.

\begin{figure}[t]
\centerline{\includegraphics[width=0.85\textwidth]{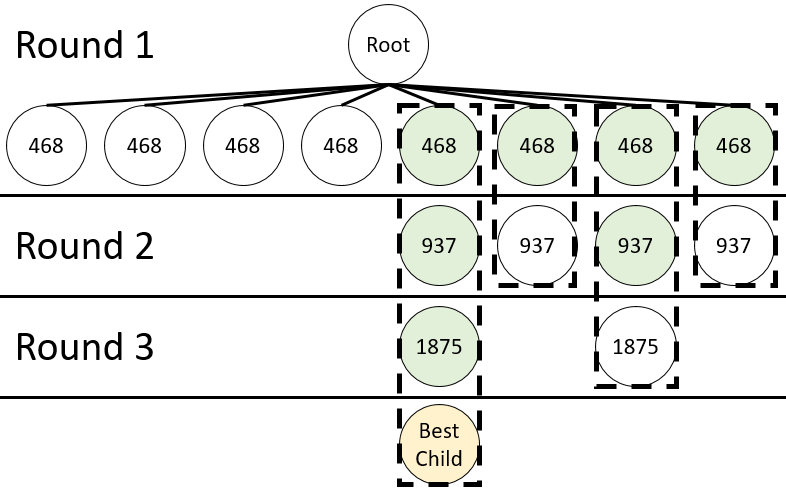}}
\caption{Example of one complete run of Sequential Halving with 15,000 iterations. Only the root node and its immediate children are depicted. Circles in the same vertical bar depict the same node, in different rounds of the algorithm. The best half of nodes in each round are colored. Numbers inside the nodes denote the total number of iterations allocated to that node up to and including the corresponding round.}
\label{fig:sh diagram}
\end{figure}

Sequential Halving (SH) \cite{Karnin_2013_Almost} is a different MAB algorithm, designed to optimize for simple regret rather than cumulative regret. It assumes prior knowledge of the budget of arm pulls (or time steps) $T$, as well as the fixed number of arms $K$. It calculates how many times the number of arms can be halved before narrowing down to a single arm as $B = \lceil \log_2 K \rceil$, and operates in $B$ separate phases. The total budget $T$ gets spread out evenly over these $B$ phases. Within every phase, the algorithm distributes the number of arm pulls uniformly over all remaining arms (in the first phase, this is the full set of all $K$ arms). After every phase, the worst-performing half of all arms get discarded. \refalgorithm{alg:SH} provides pseudocode for SH, and \reffigure{fig:sh diagram} gives a visual depiction of the algorithm.

\begin{algorithm}[t]
\caption{Sequential Halving}\label{alg:SH}
\textbf{Input} total budget $T$, $K$ arms
\begin{algorithmic}[1]
\State $S_1 \gets \left \{1,...,K\right\}, B \gets \left \lceil \log_{2}K \right \rceil$
\For{$k = 1 \dots B$}
\State Sample each arm $i \in S_k$, $n_k = \frac{T}{\left | S_k \right | \times B}$ times
\State Update the average reward for each arm relative to the found rewards
\State $S_{k+1}\gets\textrm{the}\left\lceil \left | S_k \right | /2\right \rceil \textrm{ best arms from } S_k $
\EndFor
\State \textbf{return} the sole remaining arm of $S_{B+1}$
\end{algorithmic}
\label{algo:sh}
\end{algorithm} 

Outside of affecting rewards that are collected and information that is gathered by an agent, decisions made in earlier time steps have no effect on later time steps in MAB problems. In contrast, games are examples of \textit{sequential} decision-making problems, where actions taken (moves played) in early states affect which states the agent transitions into. Monte-Carlo Tree Search (MCTS) \cite{Kocsis_2006_Bandit,Coulom_2007_MCTS} can handle this sequential nature by gradually building up and traversing through a search tree, with nodes representing different states, connected by the actions that lead to transitions between the states. When traversing the tree, MCTS views the problem of selecting a branch from each node it reaches as a MAB problem. After running many iterations, it will use the information it has collected from simulations to make a final decision as to which move to play. Because only this final move selection truly matters---any choices made for tree traversal during the search itself only result in fictitious rewards---it can be argued that algorithms that optimize for simple regret (such as SH) are more suitable than ones that optimize for cumulative regret (such as UCB1) in the root node \cite{Feldman_2014_Simple,Pepels_2014_Minimizing,Cazenave_2015_SHOT}. Outside of the root node, the argument in favor of simple regret minimization still holds due to the minimax structure of zero-sum games. However, simultaneously there is an argument to be made in favor of cumulative regret minimization, as this leads to a more focused best-first search that wastes less time on seemingly uninteresting parts of the search space---in particular for small search budgets. Hybrid MCTS (H-MCTS) \cite{Pepels_2014_Minimizing} therefore introduces a parameter that dynamically adjusts from using SH in and close to the root, to using UCB1 further down the tree, depending on the available budget. As a simplified version of this, we use H-MCTS with SH only in the root node, and UCB1 in all other nodes, as a baseline in our experiments.

\section{Time-Based and Anytime Sequential Halving} \label{sec:TimedBasedAnytimeSH}

As an initial step towards an anytime variant of SH, we consider \textit{Time-based SH}. Standard SH requires a budget of iterations to be known in advance, because it spreads this number of iterations evenly over all of its phases. If every iteration takes approximately the same amount of time,
a simple but effective algorithm with a time budget could simply divide the time budget over the phases instead of the iteration budget. This is a trivial variant of SH, but not one that we have seen in prior literature. It does not have the anytime property, as it still expects the time budget to be predetermined.

\begin{figure*}[t]
    \centering
    \begin{minipage}[b]{0.49\linewidth}
        \centering
        \includegraphics[width=\linewidth]{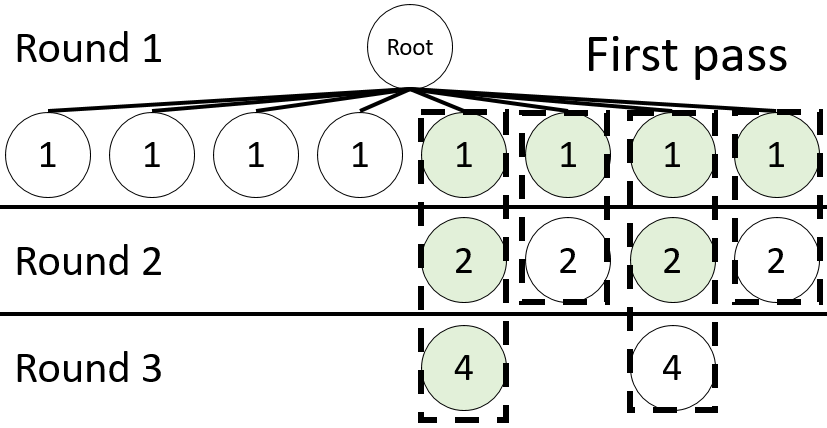}
        \label{fig:figure1}
    \end{minipage}
    \begin{minipage}[b]{0.49\linewidth}
        \centering
        \includegraphics[width=\linewidth]{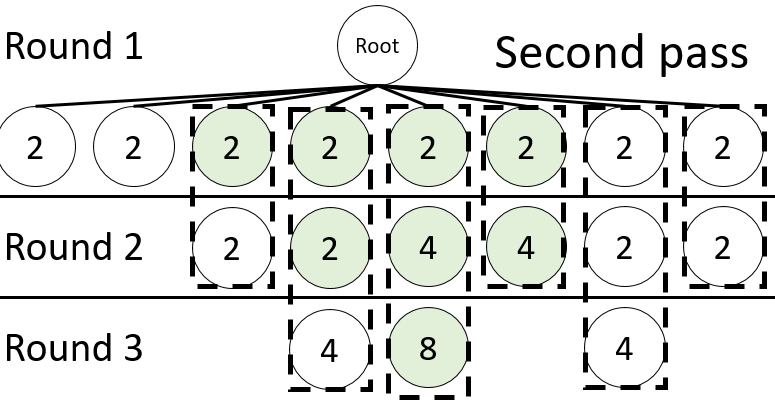}
        \label{fig:figure2}
    \end{minipage}
    \caption{Visualization of two passes made by Anytime Sequential Halving. Only the root node and its immediate children are depicted. Circles in the same vertical bar depict the same node, in different rounds of the algorithm. The best half of nodes in each round of each pass are colored. Numbers inside the nodes denote the total number of iterations allocated to that node up to and including the corresponding pass and round.}
    \label{fig:Anytimesh}
\end{figure*}

We propose \textit{Anytime SH} as an anytime variant of the algorithm, which works as follows. It starts operating like the standard SH, albeit with an assumed budget of iterations that is only the bare minimum required for a valid, ``complete'' run of SH. In the first phase it allocates exactly one iteration to each arm. Then it halves the arms, allocating two additional iterations to each remaining arm, and so on, until only a single arm remains. We refer to this as one pass of the algorithm. After the first pass, if there is still time remaining, the algorithm resets back to considering the full set of arms again, and repeats the same process. If the ordering of arms' empirically observed mean rewards never changes (for example, if every arm $i$ deterministically gives the same reward $\mu_i$ on each pull, with no randomness), this algorithm distributes its pulls over the arms in the same way as SH would have done in hindsight (if the iteration budget had been known in advance). If the ordering of arms based on their empirically observed mean rewards can vary over a run of the algorithm, it is possible to end up with different allocations of iterations than in the original SH. \refalgorithm{algo:anytime} provides pseudocode for the algorithm, and \reffigure{fig:Anytimesh} visualises two passes of the algorithm.

\begin{algorithm}[t]
\caption{Anytime Sequential Halving}\label{alg:anytime}
\textbf{Input} $K$ arms
\begin{algorithmic}[1]
\State $S \gets \left \{1,...,K\right\}$
\State $T \gets S$
\State $N \gets 1$
\Repeat
\State Every arm in $T$ is sampled $N$ times.
\State Update the average reward for each arm.
\State $T \gets$ the $\lceil |T| / 2 \rceil$ best arms from $T$
\State $N \gets N\times2$
\If{$\left | T \right | = 1$}
\State $T \gets S$
\State $N \gets 1$
\EndIf
\Until{stopping condition}
\State \textbf{return} The highest-reward arm from $S$
\end{algorithmic}
\label{algo:anytime}
\end{algorithm}

\section{Experiments}\label{Experiments}

This section describes the experiments used to evaluate the novel \textit{Anytime SH} algorithm.\footnote{Source code: \url{https://github.com/dominic-sagers/Anytime-Sequential-Halving}} The algorithm is compared to appropriate baselines in synthetic MAB problems (\refsubsection{Subsec:MABExperiments}), and in a set of ten different board games (\refsubsection{Subsec:BoardGameExperiments}). In the board games, the algorithm is not used standalone, but as selection strategy in the root node for MCTS.

\subsection{Synthetic Multi-Armed Bandit Problems} \label{Subsec:MABExperiments}

For an evaluation of the raw MAB algorithms, without the added complexity of sequential decision making and combining with tree search in games, we construct a set of 100 synthetic MAB problems as follows. For every MAB problem, we sample ten different means $\mu_1, \dots, \mu_{10}$ for $K = 10$ arms from a normal distribution with mean $0$ and unit variance. Within each problem, when an arm $i$ is pulled, a reward is drawn from another normal distribution with mean $\mu_i$ and unit variance. 
The following algorithms are compared on this suite of problems:
\begin{itemize}
    \item \textbf{UCB1}: UCB1 \cite{Auer_2002_Finite} using an exploration constant $C = \sqrt{2}^{-1}$, as described in \refsection{Sec:Background}.
    \item \textbf{Base SH}: standard Sequential Halving \cite{Karnin_2013_Almost}, as described in \refsection{Sec:Background}.
    \item \textbf{Time SH}: the trivial Time-based variant of SH, as described in \refsection{sec:TimedBasedAnytimeSH}.
    \item \textbf{Anytime SH}: the novel algorithm as described in \refsection{sec:TimedBasedAnytimeSH}.
\end{itemize}

\begin{figure*}[t]
    \centering
    \includegraphics[width=.8\linewidth]{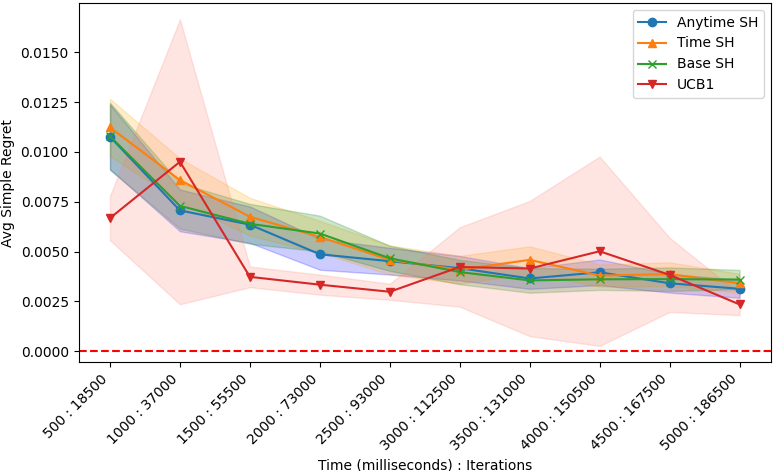} 
    \caption{Simple regret, averaged over 100 MAB problems, for four MAB algorithms.}
    \label{fig:all algos MAB}
\end{figure*}

Every algorithm was run on each MAB problem for up to 5000 milliseconds (or 186,500 iterations in the case of Base SH) using Python 3, on an i7-10700k Intel CPU. \reftable{table:mappings} lists mappings between time and iteration budgets that were empirically determined, allowing for a fair comparison between algorithms that support only certain types of budgets. For each time budget listed in \reftable{table:mappings} (or iteration budget for Base SH), the simple regret incurred by each algorithm running with that budget is recorded for each of the 100 problems, and averaged over those 100 problems. \reffigure{fig:all algos MAB} depicts $95\%$ confidence intervals for these averaged simple regret measures.

\begin{table}[t]
\caption{Mappings between time and iteration budgets for $10$-armed MAB problems.}
\centering
\begin{tabular}{@{}lrrrrrrrrrr@{}}
\toprule
\textbf{Milliseconds} & 500 & 1000 & 1500 & 2000 & 2500 & 3000 & 3500 & 4000 & 4500 & 5000 \\
\textbf{Iterations} & 18500 & 37000 & 55500 & 73000 & 93000 & 112500 & 131000 & 150500 & 167500 & 186500 \\
\bottomrule
\end{tabular}
\label{table:mappings}
\end{table}

\subsection{Board Games} \label{Subsec:BoardGameExperiments}

\begin{figure*}[t!]
    \centering
    \includegraphics[width=.45\linewidth]{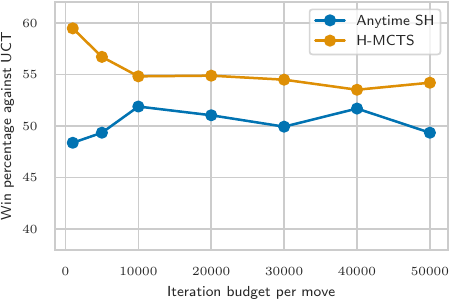} 
    \hfill
    \includegraphics[width=.45\linewidth]{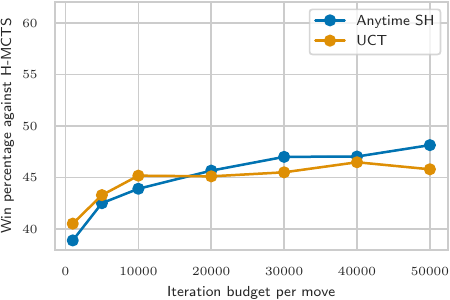}
    \caption{Win percentages averaged over all ten games against UCT (left) and H-MCTS (right). $95\%$ confidence intervals are too small to be visible.}
    \label{fig:GameSummaryPlots}
\end{figure*}

The second experiment evaluates the performance of Anytime SH, when used as selection strategy in the root node of MCTS, in a set of ten different board games (listed in \reftable{Table:GameResults}). Three different agents are considered in this experiment:
\begin{itemize}
    \item \textbf{UCT}: a standard implementation of MCTS, using UCB1 with $C=\sqrt{2}$ as selection strategy, adding one new node to the tree per iteration, and running uniformly random play-outs. 
    \item \textbf{H-MCTS}: a simplified version of Hybrid MCTS \cite{Pepels_2014_Minimizing}, using SH as selection strategy in the root node, but otherwise running as described above for UCT.
    \item \textbf{Anytime SH}: uses anytime SH as selection strategy for the root node, but otherwise runs as described above for UCT.
\end{itemize}
Each of these agents played 150 times (75 as first and 75 as second player) against each of the other two agents, across ten different board games. For each matchup, seven different MCTS iteration budgets per move were used, ranging from 1000 to 50,000. We used iteration budgets rather than time budgets, because H-MCTS can only function with the former, whereas the other two algorithms can handle either type of budget. \reftable{Table:GameResults} lists $95\%$ Agresti-Coull confidence intervals for win percentages, with any draws being counted as half-wins. All listed games were run using their default settings in the Ludii general game playing system \cite{Piette_2020_Ludii}, except for Atari Go and Clobber being played on boards of size 9$\times$9 and 7$\times$7, respectively. In addition to the table with results per game, \reffigure{fig:GameSummaryPlots} depicts the win percentages when averaging over all ten games.

\begin{table}[t!]
\setlength{\tabcolsep}{3pt}
\caption{$95\%$ Agresti-Coull confidence intervals for win percentages of row agents against UCT (top half) or H-MCTS (bottom half) in ten different board games.}
\centering
\resizebox{\textwidth}{!}{%
\begin{tabular}{@{}lrrrrrrr@{}}
\toprule
\textbf{MCTS iterations per move:} & 1000 & 5000 & 10,000 & 20,000 & 30,000 & 40,000 & 50,000 \\
\midrule
& \multicolumn{7}{c}{Win percentages \textbf{against UCT} (95\% Agresti-Coull confidence intervals)} \\
\cmidrule(lr){2-8}
\textbf{Amazons} & & & & & & & \\
\hspace{3mm}H-MCTS & $46.10 \pm 7.88$ & $56.50 \pm 7.83$ & $55.85 \pm 7.85$ & $51.95 \pm 7.90$ & $51.30 \pm 7.90$ & $46.75 \pm 7.88$ & $55.85 \pm 7.85$ \\
\hspace{3mm}Anytime SH & $51.95 \pm 7.90$ & $52.60 \pm 7.89$ & $50.00 \pm 7.90$ & $47.40 \pm 7.89$ & $46.10 \pm 7.88$ & $48.70 \pm 7.90$ & $48.05 \pm 7.90$ \\
\textbf{Atari Go (9$\times$9)} & & & & & & & \\
\hspace{3mm}H-MCTS & $64.30 \pm 7.57$ & $56.50 \pm 7.83$ & $61.70 \pm 7.68$ & $64.30 \pm 7.57$ & $65.60 \pm 7.51$ & $59.10 \pm 7.77$ & $52.60 \pm 7.89$ \\
\hspace{3mm}Anytime SH & $44.80 \pm 7.86$ & $46.75 \pm 7.88$ & $61.05 \pm 7.71$ & $53.25 \pm 7.88$ & $51.30 \pm 7.90$ & $49.35 \pm 7.90$ & $45.45 \pm 7.87$ \\
\textbf{Breakthrough} & & & & & & & \\
\hspace{3mm}H-MCTS & $47.40 \pm 7.89$ & $51.30 \pm 7.90$ & $51.30 \pm 7.90$ & $52.60 \pm 7.89$ & $58.45 \pm 7.79$ & $48.70 \pm 7.90$ & $55.20 \pm 7.86$ \\
\hspace{3mm}Anytime SH & $51.95 \pm 7.90$ & $53.90 \pm 7.88$ & $57.80 \pm 7.80$ & $61.05 \pm 7.71$ & $55.20 \pm 7.86$ & $52.60 \pm 7.89$ & $52.60 \pm 7.89$ \\
\textbf{Clobber (7$\times$7)} & & & & & & & \\
\hspace{3mm}H-MCTS & $54.55 \pm 7.87$ & $52.60 \pm 7.89$ & $46.10 \pm 7.88$ & $56.50 \pm 7.83$ & $51.95 \pm 7.90$ & $49.35 \pm 7.90$ & $48.70 \pm 7.90$ \\
\hspace{3mm}Anytime SH & $44.15 \pm 7.85$ & $51.95 \pm 7.90$ & $49.35 \pm 7.90$ & $51.95 \pm 7.90$ & $47.40 \pm 7.89$ & $51.30 \pm 7.90$ & $45.45 \pm 7.87$ \\
\textbf{Gomoku} & & & & & & & \\
\hspace{3mm}H-MCTS & $75.35 \pm 6.81$ & $78.60 \pm 6.48$ & $80.55 \pm 6.25$ & $76.65 \pm 6.69$ & $79.25 \pm 6.41$ & $72.75 \pm 7.04$ & $74.05 \pm 6.93$ \\
\hspace{3mm}Anytime SH & $55.20 \pm 7.86$ & $55.20 \pm 7.86$ & $50.00 \pm 7.90$ & $45.45 \pm 7.87$ & $45.45 \pm 7.87$ & $47.40 \pm 7.89$ & $59.10 \pm 7.77$ \\
\textbf{Hex} & & & & & & & \\
\hspace{3mm}H-MCTS & $79.25 \pm 6.41$ & $63.00 \pm 7.63$ & $57.80 \pm 7.80$ & $53.25 \pm 7.88$ & $48.70 \pm 7.90$ & $59.10 \pm 7.77$ & $54.55 \pm 7.87$ \\
\hspace{3mm}Anytime SH & $32.45 \pm 7.40$ & $40.90 \pm 7.77$ & $52.60 \pm 7.89$ & $46.10 \pm 7.88$ & $53.90 \pm 7.88$ & $54.55 \pm 7.87$ & $48.05 \pm 7.90$ \\
\textbf{Pentalath} & & & & & & & \\
\hspace{3mm}H-MCTS & $62.35 \pm 7.66$ & $59.75 \pm 7.75$ & $56.50 \pm 7.83$ & $50.00 \pm 7.90$ & $48.70 \pm 7.90$ & $53.90 \pm 7.88$ & $52.60 \pm 7.89$ \\
\hspace{3mm}Anytime SH & $52.60 \pm 7.89$ & $44.80 \pm 7.86$ & $51.30 \pm 7.90$ & $51.30 \pm 7.90$ & $53.25 \pm 7.88$ & $57.80 \pm 7.80$ & $47.40 \pm 7.89$ \\
\textbf{Reversi} & & & & & & & \\
\hspace{3mm}H-MCTS & $50.98 \pm 7.90$ & $51.95 \pm 7.90$ & $51.30 \pm 7.90$ & $53.25 \pm 7.88$ & $46.10 \pm 7.88$ & $45.77 \pm 7.87$ & $49.35 \pm 7.90$ \\
\hspace{3mm}Anytime SH & $45.45 \pm 7.87$ & $53.25 \pm 7.88$ & $52.28 \pm 7.89$ & $55.85 \pm 7.85$ & $51.63 \pm 7.90$ & $51.95 \pm 7.90$ & $49.35 \pm 7.90$ \\
\textbf{Tablut} & & & & & & & \\
\hspace{3mm}H-MCTS & $57.15 \pm 7.82$ & $50.65 \pm 7.90$ & $50.00 \pm 7.90$ & $50.00 \pm 7.90$ & $49.35 \pm 7.90$ & $50.00 \pm 7.90$ & $50.65 \pm 7.90$ \\
\hspace{3mm}Anytime SH & $48.05 \pm 7.90$ & $50.00 \pm 7.90$ & $50.65 \pm 7.90$ & $50.00 \pm 7.90$ & $51.95 \pm 7.90$ & $50.00 \pm 7.90$ & $49.35 \pm 7.90$ \\
\textbf{Yavalath} & & & & & & & \\
\hspace{3mm}H-MCTS & $57.15 \pm 7.82$ & $46.10 \pm 7.88$ & $37.00 \pm 7.63$ & $40.25 \pm 7.75$ & $45.45 \pm 7.87$ & $49.67 \pm 7.90$ & $48.37 \pm 7.90$ \\
\hspace{3mm}Anytime SH & $57.15 \pm 7.82$ & $44.15 \pm 7.85$ & $43.82 \pm 7.84$ & $48.05 \pm 7.90$ & $43.17 \pm 7.83$ & $53.25 \pm 7.88$ & $48.70 \pm 7.90$ \\
\midrule
& \multicolumn{7}{c}{Win percentages \textbf{against H-MCTS} (95\% Agresti-Coull confidence intervals)} \\
\cmidrule(lr){2-8}
\textbf{Amazons} & & & & & & & \\
\hspace{3mm}UCT & $53.90 \pm 7.88$ & $43.50 \pm 7.83$ & $44.15 \pm 7.85$ & $48.05 \pm 7.90$ & $48.70 \pm 7.90$ & $53.25 \pm 7.88$ & $44.15 \pm 7.85$ \\
\hspace{3mm}Anytime SH & $38.95 \pm 7.71$ & $37.00 \pm 7.63$ & $41.55 \pm 7.79$ & $51.95 \pm 7.90$ & $40.90 \pm 7.77$ & $44.15 \pm 7.85$ & $40.25 \pm 7.75$ \\
\textbf{Atari Go (9$\times$9)} & & & & & & & \\
\hspace{3mm}UCT & $35.70 \pm 7.57$ & $43.50 \pm 7.83$ & $38.30 \pm 7.68$ & $35.70 \pm 7.57$ & $34.40 \pm 7.51$ & $40.90 \pm 7.77$ & $47.40 \pm 7.89$ \\
\hspace{3mm}Anytime SH & $33.75 \pm 7.47$ & $42.20 \pm 7.80$ & $33.75 \pm 7.47$ & $45.45 \pm 7.87$ & $41.55 \pm 7.79$ & $40.25 \pm 7.75$ & $50.00 \pm 7.90$ \\
\textbf{Breakthrough} & & & & & & & \\
\hspace{3mm}UCT & $52.60 \pm 7.89$ & $48.70 \pm 7.90$ & $48.70 \pm 7.90$ & $47.40 \pm 7.89$ & $41.55 \pm 7.79$ & $51.30 \pm 7.90$ & $44.80 \pm 7.86$ \\
\hspace{3mm}Anytime SH & $50.65 \pm 7.90$ & $52.60 \pm 7.89$ & $53.90 \pm 7.88$ & $46.75 \pm 7.88$ & $50.65 \pm 7.90$ & $56.50 \pm 7.83$ & $50.65 \pm 7.90$ \\
\textbf{Clobber (7$\times$7)} & & & & & & & \\
\hspace{3mm}UCT & $45.45 \pm 7.87$ & $47.40 \pm 7.89$ & $53.90 \pm 7.88$ & $43.50 \pm 7.83$ & $48.05 \pm 7.90$ & $50.65 \pm 7.90$ & $51.30 \pm 7.90$ \\
\hspace{3mm}Anytime SH & $54.55 \pm 7.87$ & $46.75 \pm 7.88$ & $47.40 \pm 7.89$ & $51.95 \pm 7.90$ & $50.65 \pm 7.90$ & $53.25 \pm 7.88$ & $49.35 \pm 7.90$ \\
\textbf{Gomoku} & & & & & & & \\
\hspace{3mm}UCT & $24.65 \pm 6.81$ & $21.40 \pm 6.48$ & $19.45 \pm 6.25$ & $23.35 \pm 6.69$ & $20.75 \pm 6.41$ & $27.25 \pm 7.04$ & $25.95 \pm 6.93$ \\
\hspace{3mm}Anytime SH & $32.45 \pm 7.40$ & $31.80 \pm 7.36$ & $18.15 \pm 6.09$ & $10.35 \pm 4.81$ & $25.30 \pm 6.87$ & $18.15 \pm 6.09$ & $26.60 \pm 6.98$ \\
\textbf{Hex} & & & & & & & \\
\hspace{3mm}UCT & $20.75 \pm 6.41$ & $37.00 \pm 7.63$ & $42.20 \pm 7.80$ & $46.75 \pm 7.88$ & $51.30 \pm 7.90$ & $40.90 \pm 7.77$ & $45.45 \pm 7.87$ \\
\hspace{3mm}Anytime SH & $12.30 \pm 5.19$ & $22.70 \pm 6.62$ & $38.30 \pm 7.68$ & $45.45 \pm 7.87$ & $51.95 \pm 7.90$ & $50.00 \pm 7.90$ & $48.70 \pm 7.90$ \\
\textbf{Pentalath} & & & & & & & \\
\hspace{3mm}UCT & $37.65 \pm 7.66$ & $40.25 \pm 7.75$ & $43.50 \pm 7.83$ & $50.00 \pm 7.90$ & $51.30 \pm 7.90$ & $46.10 \pm 7.88$ & $47.40 \pm 7.89$ \\
\hspace{3mm}Anytime SH & $24.65 \pm 6.81$ & $35.70 \pm 7.57$ & $44.80 \pm 7.86$ & $50.00 \pm 7.90$ & $43.50 \pm 7.83$ & $54.55 \pm 7.87$ & $55.20 \pm 7.86$ \\
\textbf{Reversi} & & & & & & & \\
\hspace{3mm}UCT & $49.02 \pm 7.90$ & $48.05 \pm 7.90$ & $48.70 \pm 7.90$ & $46.75 \pm 7.88$ & $53.90 \pm 7.88$ & $54.23 \pm 7.87$ & $50.65 \pm 7.90$ \\
\hspace{3mm}Anytime SH & $50.98 \pm 7.90$ & $51.95 \pm 7.90$ & $54.23 \pm 7.87$ & $48.05 \pm 7.90$ & $64.95 \pm 7.54$ & $54.23 \pm 7.87$ & $59.43 \pm 7.76$ \\
\textbf{Tablut} & & & & & & & \\
\hspace{3mm}UCT & $42.85 \pm 7.82$ & $49.35 \pm 7.90$ & $50.00 \pm 7.90$ & $50.00 \pm 7.90$ & $50.65 \pm 7.90$ & $50.00 \pm 7.90$ & $49.35 \pm 7.90$ \\
\hspace{3mm}Anytime SH & $43.50 \pm 7.83$ & $47.40 \pm 7.89$ & $48.70 \pm 7.90$ & $49.35 \pm 7.90$ & $48.05 \pm 7.90$ & $48.05 \pm 7.90$ & $47.40 \pm 7.89$ \\
\textbf{Yavalath} & & & & & & & \\
\hspace{3mm}UCT & $42.85 \pm 7.82$ & $53.90 \pm 7.88$ & $63.00 \pm 7.63$ & $59.75 \pm 7.75$ & $54.55 \pm 7.87$ & $50.33 \pm 7.90$ & $51.63 \pm 7.90$ \\
\hspace{3mm}Anytime SH & $47.40 \pm 7.89$ & $57.15 \pm 7.82$ & $58.45 \pm 7.79$ & $57.48 \pm 7.81$ & $52.60 \pm 7.89$ & $51.30 \pm 7.90$ & $53.90 \pm 7.88$ \\
\bottomrule
\end{tabular}}
\label{Table:GameResults}
\end{table}

\section{Discussion}\label{Discussion}
The results for the MAB problems in \reffigure{fig:all algos MAB} show that Anytime SH performs on par with the naive time-based SH and the standard SH in terms of simple regret. In this setting, adjusting the algorithm to introduce the anytime property does not harm performance. UCB1 appears to have better performance for some budgets, which is somewhat surprising as this algorithm is not necessarily designed to optimize simple regret. However, for many budgets, UCB1 has substantially greater variance in its performance level than the three SH-based algorithms do.

When averaging results over all ten board games (\reffigure{fig:GameSummaryPlots}), we find that (i) H-MCTS (which is not an anytime algorithm) tends to perform better than the other two algorithms, in particular for low iteration budgets, and (ii) anytime SH appears to be possibly weaker than UCT by a slight margin for low iteration budgets, but evenly-matched or slightly stronger for medium and large iteration budgets. At the level of individual games, these trends can differ (see \reftable{Table:GameResults}). The results in the game of Gomoku stand out in particular, with H-MCTS having particularly high win percentages of around $75\%$ against UCT for all budgets, and ranging between approximately $70\%$ and $90\%$ against Anytime SH. While a sample size of ten games (with 150 plays per game, per pair of agents) is no less than is customary in the literature, it is still small enough for a single game with relatively extreme results such as Gomoku to have an outsized effect on a plot that averages over the games like \reffigure{fig:GameSummaryPlots}. Especially in the direct matchup against H-MCTS (the left subfigure), Anytime SH would be substantially closer to H-MCTS if the results for Gomoku were excluded (and, plausibly also if the experiment were extended with more other games, if Gomoku is an outlier).

\section{Conclusion and Future Work}\label{Conclusion}

This paper proposed Anytime Sequential Halving: an algorithm designed in an effort to approximate the behavior of the standard Sequential Halving (SH) algorithm, whilst having the \textit{anytime} property. Experiments in a set of synthetic Multi-Armed Bandit problems show that Anytime SH performs on par in terms of simple regret with the standard SH, as well as a trivial time-based variant of SH, which can handle time-based budgets (but is not truly anytime). While the performance in terms of simple regret remains unchanged, Anytime SH does bring the benefit of having the anytime property. When used as selection strategy for the root node in Monte-Carlo Tree Search (MCTS) for game playing, Anytime SH appears to perform slightly below the Hybrid MCTS baseline (which has the downside of not being an anytime algorithm). Its performance level is competitive with UCT, appearing to be possibly slightly worse for low search budgets, but slightly better for higher budgets.

We see potential for ample future work to further improve Anytime SH. The Anytime SH algorithm as proposed in this paper was kept simple and straightforward. We expect that there is likely room to improve the algorithm, in particular by more carefully investigating how it should behave in situations where its ordering of arms changes between different passes of the algorithm (a situation that cannot occur in the standard SH). When Anytime SH ``changes its mind'' about ordering of any pair of arms, one of them will have had fewer iterations so far than it should have had in hindsight (fewer iterations than the standard SH would have allocated if the current total number of iterations were the full budget). In our implementation, we ignore this, but it is worth investigating if correcting for this would be worthwhile. This could potentially make the algorithm more adaptive to the level of complexity of a problem than the standard SH is, which always halves sets of arms at specific intervals regardless of how many arms are close or far from each other in terms of expected rewards. Furthermore, it would be interesting to investigate how the algorithm interacts with different values of hyperparameters, such as the exploration constant still used for UCB1 in nodes below the root node in MCTS.



\begin{credits}

\subsubsection{\discintname}
The authors have no competing interests to declare that are relevant to the content of this article.
\end{credits}
%
%
%
\bibliographystyle{splncs04}
\bibliography{Dennis-Soemers-BibABBREV}
\end{document}